\documentclass[sigconf]{acmart}
\AtBeginDocument{%
  \providecommand\BibTeX{{%
    \normalfont B\kern-0.5em{\scshape i\kern-0.25em b}\kern-0.8em\TeX}}}

\copyrightyear{2023}
\acmYear{2023}
\setcopyright{acmlicensed}\acmConference[MM '23]{Proceedings of the 31st ACM International Conference on Multimedia}{October 29-November 3, 2023}{Ottawa, ON, Canada}
\acmBooktitle{Proceedings of the 31st ACM International Conference on Multimedia (MM '23), October 29-November 3, 2023, Ottawa, ON, Canada}
\acmPrice{15.00}
\acmDOI{10.1145/3581783.3611832}
\acmISBN{979-8-4007-0108-5/23/10}

\renewcommand\footnotetextcopyrightpermission[1]{}   %remove ACM copyright format
\acmSubmissionID{601}

\usepackage{bm}

\begin{document}

\title{Learning Comprehensive Representations with Richer Self for Text-to-Image Person Re-Identification}

% \author{Shuanglin~Yan$^1$, Neng~Dong$^1$, Jun~Liu$^2$, Liyan~Zhang$^{3}$, Jinhui~Tang$^1$}
% \affiliation{$^1$School of Computer Science and Engineering, Nanjing University of Science and Technology\country{China}}
% \affiliation{$^2$Information Systems Technology and Design Pillar, Singapore University of Technology and Design\country{Singapore}}
% \affiliation{$^3$School of Computer Science and Technology, Nanjing University of Aeronautics and Astronautics\country{China}}
% \email{{shuanglinyan, neng.dong, jinhuitang}@njust.edu.cn, jun_liu@sutd.edu.sg, zhangliyan@nuaa.edu.cn}

\author{Shuanglin Yan}
\affiliation{%
  \institution{Nanjing University of Science and Technology}
  \state{Nanjing}
  \country{China}
}
\email{shuanglinyan@njust.edu.cn}

\author{Neng Dong}
\affiliation{%
  \institution{Nanjing University of Science and Technology}
  \state{Nanjing}
  \country{China}}
\email{neng.dong@njust.edu.cn}

\author{Jun Liu}
\affiliation{%
  \institution{Singapore University of Technology and Design}
  \streetaddress{8 Somapah Rd}
  \country{Singapore}}
\email{jun_liu@sutd.edu.sg}

\author{Liyan Zhang}\authornote{Corresponding author.}
\affiliation{%
  \institution{Nanjing University of Aeronautics and Astronautics}
  \state{Nanjing}
  \country{China}}
\email{zhangliyan@nuaa.edu.cn}
  
\author{Jinhui Tang}
\affiliation{%
  \institution{Nanjing University of Science and Technology}
  \state{Nanjing}
  \country{China}}
\email{jinhuitang@njust.edu.cn}

\renewcommand{\shortauthors}{Shuanglin Yan, Neng Dong, Jun Liu, Liyan Zhang, \& Jinhui Tang}

\begin{abstract}
Text-to-image person re-identification (TIReID) retrieves pedestrian images of the same identity based on a query text. However, existing methods for TIReID typically treat it as a one-to-one image-text matching problem, only focusing on the relationship between image-text pairs within a view. The \textbf{many-to-many matching} between image-text pairs across views under the same identity is not taken into account, which is one of the main reasons for the poor performance of existing methods. To this end, we propose a simple yet effective framework, called \textbf{LCR$^2$S}, for modeling many-to-many correspondences of the same identity by learning comprehensive representations for both modalities from a novel perspective. We construct a support set for each image (text) by using other images (texts) under the same identity and design a multi-head attentional fusion module to fuse the image (text) and its support set. The resulting enriched image and text features fuse information from multiple views, which are aligned to train a "richer" TIReID model with many-to-many correspondences. Since the support set is unavailable during inference, we propose to distill the knowledge learned by the "richer" model into a lightweight model for inference with a single image/text as input. The lightweight model focus on semantic association and reasoning of multi-view information, which can generate a comprehensive representation containing multi-view information with only a single-view input to perform accurate text-to-image retrieval during inference. In particular, we use the intra-modal features and inter-modal semantic relations of the "richer" model to supervise the lightweight model to inherit its powerful capability. Extensive experiments demonstrate the effectiveness of LCR$^2$S, and it also achieves \textbf{new state-of-the-art performance} on three popular TIReID datasets.
\end{abstract}
% \textcolor{red}{Insight, why m2m and kd effective.}
% <concept_id>10002951.10003317.10003338.10003346</concept_id>
% <concept_desc>Information systems~Top-k retrieval in databases</concept_desc>

% \ccsdesc[500]{Information systems~Top-k retrieval in databases}


\maketitle

\begin{figure}[t!]
\setlength{\abovecaptionskip}{0.2cm}
\centering
\includegraphics[width=\linewidth]{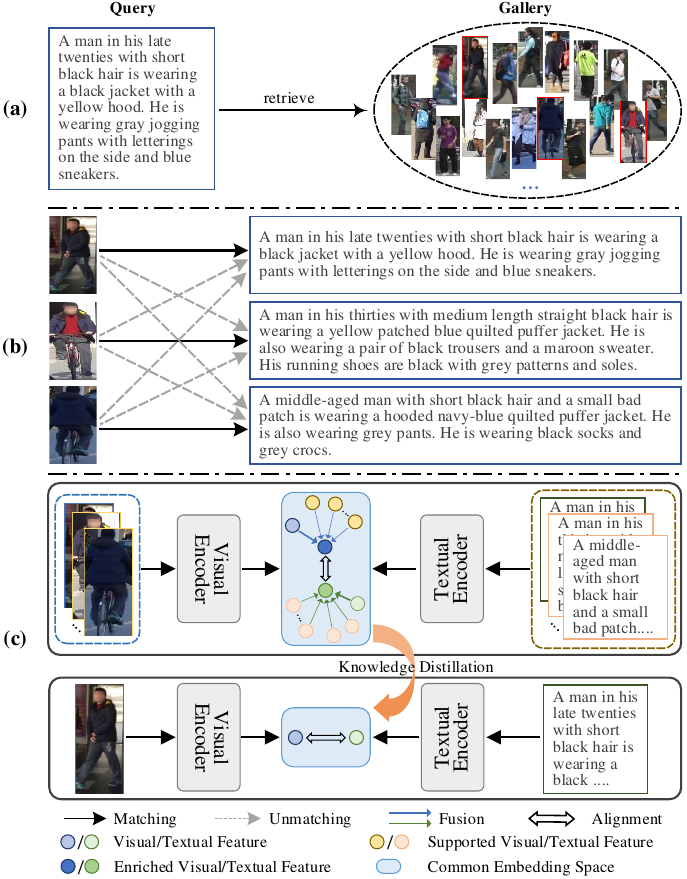}
\caption{The motivation for LCR$^2$S. (a) TIReID is to find images of the same identity across multiple views using a single-view text. (b) Existing methods only consider the one-to-one matching of each image-text pair within a view, ignoring many-to-many matching between images and texts under the same identity. (c) We explore many-to-many correspondences by aligning the enriched image and text features that fuse information from multiple views, and distill the "richer" knowledge into a basic dual network for inference.}
\label{Fig:1}
\vspace{-0.2cm}
\end{figure}

\section{Introduction}
Person Re-identification (ReID) has gained popularity as a means of retrieving pedestrian images with the same identity as the given query across cameras. However, most existing ReID methods focus on image-to-image retrieval scenarios~\cite{PIE, AIESL, dong2023erasing, cdvr, shen2023git}, which may fail when the target pedestrian's image is not available under a certain camera. In this paper, we focus on the text-to-image retrieval scenario, i.e. text-to-image ReID (TIReID), which has image-text pairs captured from multiple views for training, while only a single-view text is used to retrieve images of the same identity from a multi-view image gallery during inference (as shown in Figure~\ref{Fig:1} (a)). TIReID remains a challenging task since images and texts have different semantical descriptions which result in a modality gap. 

The general procedure for TIReID involves encoding images and texts through a visual encoder and a textual encoder, then projecting them into a common embedding space for modality alignment. The major challenge is how to align the data pairs from the two modalities. Typically, there are two popular types of methods to align image-text pairs. One type is global-level methods~\cite{CMPM, CMKA, LapsCore, RKT}, which try to learn modality-shared global features for two modalities in the embedding space. However, the significant modality gap makes these methods difficult to align images and texts at a global level. The other type is local-level methods~\cite{NAFS, SSAN, tipcb, lgur}, which focus on mining modality-specific local features and multi-level fine-grained alignment. Local-level methods have proven to be highly effective in modality alignment and are currently the dominant method for TIReID. However, existing methods treat TIReID as a general image-text matching problem, only considering the one-to-one matching of each image and paired text within a view. As illustrated in Figure~\ref{Fig:1} (b), different from conventional image-text matching, TIReID has multiple image-text pairs (each row) from multiple views under the same identity, which involves \textbf{many-to-many matching} (different rows) between images and texts under the same identity across views, rather than just a one-to-one matching (each row) between a single image and paired text within a view. Thus, an appropriate solution would be to well-match each text with multiple images of the same identity and vice versa.

The simplest approach to addressing the problem is to minimize the distance between images and texts that correspond to the same identity in the joint embedding space. However, due to variations in viewpoint and language usage among different individuals, images/texts of the same pedestrian can be highly diverse. Directly matching across different views could potentially disrupt the intrinsic correspondence between the text and its corresponding image within a view, leading to significant performance degradation (see Figure~\ref{Fig:a} (contrastive loss)). Moreover, it is not feasible to match diverse images from multiple views of the same identity with only a single-view text. In the paper, we propose a novel approach to address this issue by enriching each text (image) with multiple additional texts (images) from different views of the same identity, as illustrated in Figure~\ref{Fig:1} (c). By aligning the enriched images and texts, we are able to indirectly achieve many-to-many alignment between images and texts under the same identity. However, this method has a limitation in that it requires access to additional images and texts of the same identity, which is not available during inference. Thus, we introduce knowledge distillation to train a simple and lightweight model that can perform inference using only a single text or image as input.

In summary, we present a novel \textbf{L}earning \textbf{C}omprehensive \textbf{R}epre-\\sentations with \textbf{R}icher \textbf{S}elf (\textbf{LCR$^2$S}) framework to mine many-to-many correspondences between images and texts of the same identity for TIReID. The framework includes a teacher network for learning richer information with multiple texts/images of the same identity as input and a student network with a single text/image as input for inference. In the teacher network, we first construct a textual (visual) support set for each text (image) using other texts (images) under the same identity. Then we utilize a multi-head attentional fusion module to generate a richer textual (visual) representation from the text (image) and corresponding textual (visual) support set. The generated enriched text and image representations are aligned by both multi-stage and cross-stage CMPM losses in the common embedding space. The student network is a basic dual encoding network that receives a single text/image as input, which is trained with supervision from the teacher network via knowledge distillation to inherit its rich knowledge. We leverage the intra-modal features and inter-modal semantic relations of the teacher network as supervision signals to better empower the student network with the ability of multi-view semantic association and reasoning. During inference, only the student network is used.

The main contributions are as follows: (1) We propose a simple yet effective LCR$^2$S framework for TIReID that explores a novel perspective for mining many-to-many correspondences between images and texts of the same identity. To our best knowledge, we are the first to explore the effective many-to-many correspondences for TIReID and distill it into a lightweight network for efficient inference. (2) Both multi-stage and cross-stage CMPM losses are introduced to align enriched visual/textual embeddings to model many-to-many correspondences. (3) We use the intra-modal features and inter-modal relations of the teacher network to supervise the student network for knowledge transfer. (4) We conduct extensive experiments to validate the effectiveness of our LCR$^2$S, and it achieves new state-of-the-art results on three TIReID benchmark datasets.

\vspace{-0.21cm}
\section{Related Work}
\subsection{Text-to-Image Person Re-identification}
In contrast to image-based ReID~\cite{pifhsa, shen2021exploring, shen2023triplet, GLMC}, TIReID~\cite{GNA} is more challenging due to the consideration of both intra-modal and inter-modal divergences. The TIReID methods can be classified into global alignment-based and local alignment-based methods. Early works~\cite{Dual, CMPM, MCCL, CMKA} are mostly global alignment-based, which directly projects images and texts into a joint space to learn modality-shared features. For instance, Zhang \emph{et al}.~\cite{CMPM} proposed a cross-modal projection matching (CMPM) loss and a cross-modal projection classification (CMPC) loss to learn modality-shared features. And the CMPM loss has been used as a basic loss in subsequent works. Wang \emph{et al}.~\cite{MCCL} designed a mutually connected classification loss to exploit identity-level information and encourage the cross-modal classification probabilities of the same identity to be more similar. Chen \emph{et al}.~\cite{CMKA} proposed a cross-modal knowledge adaptation model to reduce the differences between modalities by using text as a guide to suppress image-specific information. The global alignment-based methods are simple and efficient, but the performance is not satisfactory.
 
The recent dominant methods are local alignment-based, which first acquires visual and textual local features, and then mines fine-grained~\cite{ZhaTST23, LBUL, TangLPT20, LiTPQT23, agpf} correspondences between them in the joint space. To obtain local features, some methods~\cite{PMA, vitaa} introduce external models to obtain image parts and text phrases. Most methods~\cite{SSAN, NAFS, tipcb} still split images and texts into multiple local parts directly. To avoid the above explicit local feature acquisition methods, Yan \emph{et al}.~\cite{MANet} proposed an implicit local alignment to learn a set of modality-shared local features. According to the local alignment strategy, local alignment-based methods can be divided into cross-modal interaction-based and interaction-free methods. Cross-modal interaction-based methods~\cite{MIA, PMA, SCAN, NAFS, DSSL} generate locally aligned features or similarity scores through the interactions between image and text local features. Jing \emph{et al}.~\cite{PMA} utilized the pose information to guide the attention of noun phrases and image regions to generate the attended region-related text representation (phrase-related visual representation) for each image region (noun phrase). These methods can achieve superior performance but require higher computational costs. To avoid complex cross-modal interactions, many cross-modal interaction-free methods~\cite{SSAN, tipcb, lgur, SRCF} learn local features for each modality independently and then align them through loss optimization in the joint space. Some lightweight models~\cite{SSAN, tipcb, saf} are proposed that achieve state-of-the-art performance without cross-modal interactions. Recently, several works proposed to leverage the rich prior knowledge of large-scale multimodal pre-trained models to improve the performance of TIReID. Yan \emph{et al}.~\cite{CFine} and Jiang \emph{et al}.~\cite{IRRA} transfer the knowledge of CLIP~\cite{clip} to TIReID in an end-to-end manner. 

However, existing methods only consider the one-to-one matching between image-text pairs within a view, ignoring the many-to-many matching between images and texts of the same identity across views. This limitation is one of the major reasons behind the suboptimal performance of TIReID. To this end, we propose a new approach that aims to learn comprehensive representations containing multi-view information for each modality and model many-to-many correspondences across views for the same identity. This novel perspective enables us to alleviate the limitations of existing methods and improve the performance of TIReID.

\vspace{-4pt}
\subsection{Knowledge Distillation}
Knowledge distillation (KD) is a well-known technique for transferring knowledge across different networks. This technology was originally proposed for model compression~\cite{KD}, that is, using a lightweight and small model (student) to imitate the output of a heavyweight and large model (teacher), so that this lightweight model inherits the capabilities of the heavyweight model. Hinton \emph{et al}.~\cite{hinton} proposed to transfer knowledge from teacher network to student network by minimizing the Kullback-Leibler divergence between classification logits produced by two networks. Bengio \emph{et al}.~\cite{bengio} transferred knowledge by directly minimizing the Mean Square Error (MSE) of the outputs of these two networks. Pork \emph{et al}.~\cite{RKD} further distilled the mutual relations of samples from teacher model to student model. The above methods~\cite{switchable} focus on learning a lightweight student model from a teacher with the same input data. Recently, some efforts~\cite{TKP, CleanerS, HG, LINAS, VKD, UMTS} have tried to learn student models with specific abilities from teacher models with different input data. Gu \emph{et al}.~\cite{TKP} made the student network with image data as input imitate the output of the teacher network with video data as input, which makes the student network the ability to model temporal knowledge~\cite{yan2018participation, yan2020higcin, 10082892}. Kiran \emph{et al}.~\cite{HG} proposed a holistic student-teacher network that matches the distributions of between- and within-class distances (DCDs) of occluded samples with that of holistic (non-occluded) samples, improving the robustness of the student network to occlusions. Wang \emph{et al}.~\cite{CleanerS} proposed to use a teacher model with cleaner knowledge to teach the student model with noisy input the ability to denoising. Inspired by these works, in this work, we try to learn a teacher model with more comprehensive and richer knowledge, and transfer this knowledge to the student network with a single input data to make it possess the ability of multi-view semantic association and reasoning.

\begin{figure}[t!]
\setlength{\abovecaptionskip}{0.01cm}
\setlength{\belowcaptionskip}{-0.2cm}
\centering
\includegraphics[width=\linewidth]{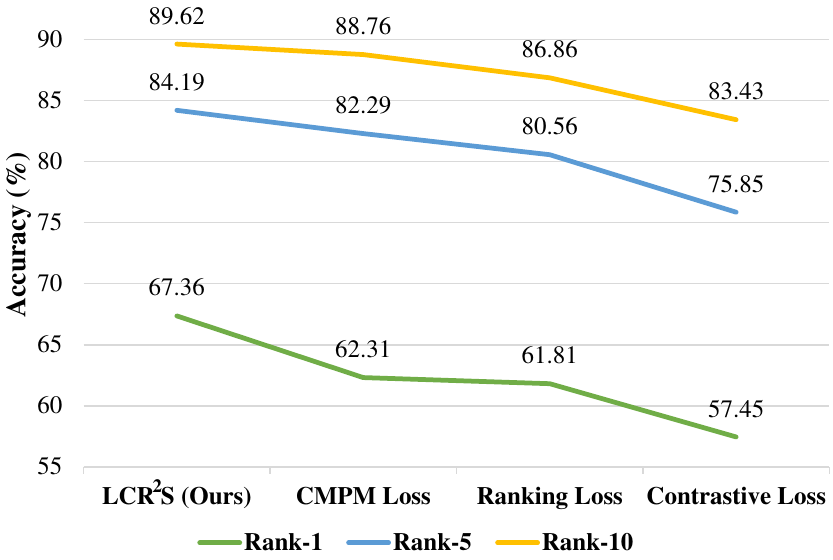}
\caption{Performance comparison of our method and Baselines trained by different alignment losses on CHUK-PEDES.}
\label{Fig:a}
\vspace{-0.3cm}
\end{figure}

\begin{figure*}
\setlength{\abovecaptionskip}{0.1cm}
\setlength{\belowcaptionskip}{-0.2cm}
\includegraphics[width=\textwidth]{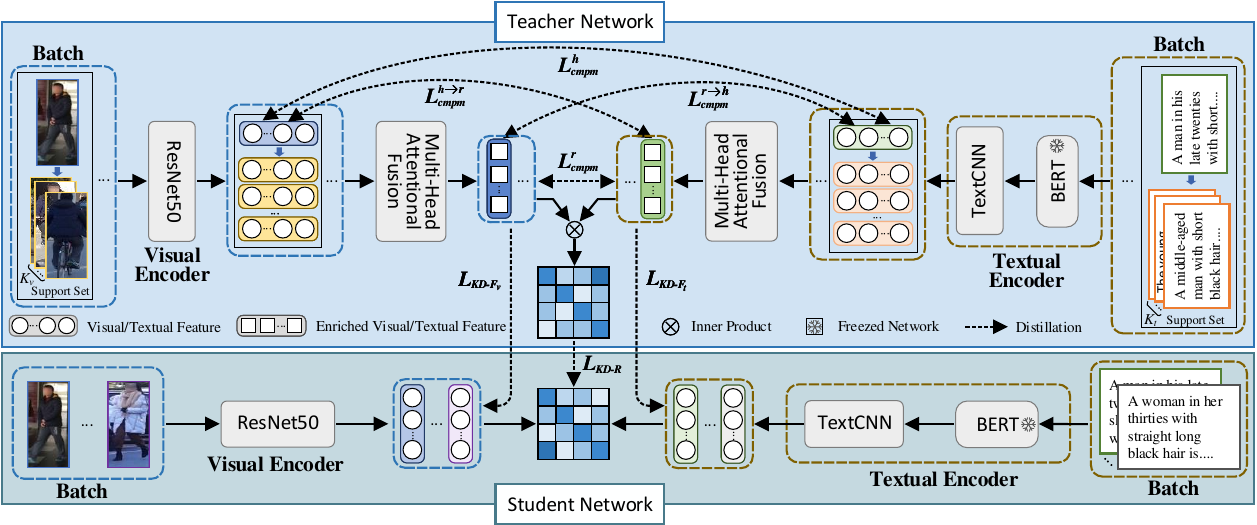}
\caption{The overall framework of LCR$^2$S. It comprises a teacher network and a student network. The teacher network with multiple texts/images and their corresponding support sets as input aims to fuse multi-view information by multi-head attentional fusion module to generate richer text/image embeddings, followed by aligning them to model many-to-many correspondences. The student network is a basic dual encoding network that takes a single text/image as input and inherits the teacher network's ability through knowledge distillation. During testing, only the student network is used for inference.}
\label{Fig:2}
\end{figure*}

\section{Methodology}
In this section, we elaborate on the implementation details of our LCR$^2$S framework, and the overview is shown in Figure~\ref{Fig:2}. In the following, we introduce cross-modal alignment objectives and identify some of their limitations in Section~\ref{sec3.1}. Section~\ref{sec3.2} and~\ref{sec3.3} elaborate on the pipelines of the teacher (Richer Self) and student (Distilling "Richer" Knowledge) models, respectively.

\subsection{Preliminaries}\label{sec3.1}
We consider a batch of $N$ paired image-text tuples $\{I_i, C_i\}_{i=1}^N$ and corresponding ground-truth label set $\{L_{i}\}_{i=1}^N$ drawn from a TIReID dataset. The goal for TIReID is to encode these data pairs into a common embedding space for cross-modal alignment. Following~\cite{tipcb}, we use ResNet50~\cite{resnet} and TextCNN~\cite{tipcb} as visual and textual encoders to extract image and text embeddings, $\bm V=\{\bm v_i\}_{i=1}^N\in \mathbb{R}^{N\times d}$ and $\bm T=\{\bm t_i\}_{i=1}^N\in \mathbb{R}^{N\times d}$, respectively. The common alignment objective functions for TIReID include cross-modality bi-directional ranking loss and cross-modal projection matching (CMPM) loss~\cite{CMPM}. The former can be expressed as follows:
\begin{equation}\
\begin{aligned}
\mathcal{L}_{rank}(\bm V, \bm T)&=\sum_{i=1}^{N}\{max(\alpha-S(\bm v_i, \bm t_i)+S(\bm v_i, \bm t_{i,n}), 0)\\
&+max(\alpha-S(\bm t_i, \bm v_i)+S(\bm t_i, \bm v_{i,n}), 0)\}
\end{aligned},
\label{eq.1}
\end{equation}
where $(\bm v_i, \bm t_{i,n}), (\bm t_i, \bm v_{i,n})$ denote the negative pairs, $S(\cdot, \cdot)$ denotes the similarity function, and $\alpha$ indicates the margin. As can be seen from Eq.~(\ref{eq.1}), the ranking loss only considers one-to-one matching between the single-view positive pair $(\bm v_i, \bm t_i)$. When there are multiple single-view image-text pairs $(\bm v_i, \bm t_i)$ and $(\bm v_j, \bm t_j)$ under the same identity, that is, $L_i=L_j$, the many-to-many matching between multiple cross-view positive pairs $(\bm v_i, \bm t_j)$, $(\bm v_j, \bm t_i)$ is not considered. Moreover, the CMPM loss can be expressed as follows:
\begin{equation}\
\begin{aligned}
p_{i,j}=\frac{exp({\bm v_i}^T\overline{\bm t}_j)}{\sum_{k=1}^{N}exp({\bm v_i}^T\overline{\bm t}_k)}~~~~~~~~s.t.~~~\overline{\bm t}_j=\frac{\bm t_j}{\|\bm t_j\|}\\
\end{aligned},
\label{eq.2}
\end{equation}
\begin{equation}\
\begin{aligned}
\mathcal{L}_{i2t}(V, T)=KL(\bm p_i\|\bm q_j)=\frac{1}{N}\sum_{i=1}^{N}\sum_{j=1}^{N}p_{i,j}log\frac{p_{i,j}}{q_{i,j}+\epsilon}\\
\end{aligned},
\label{eq.3}
\end{equation}
\begin{equation}\
\begin{aligned}
\mathcal{L}_{cmpm}(\bm V, \bm T)=\mathcal{L}_{i2t}(\bm V, \bm T)+\mathcal{L}_{t2i}(\bm V, \bm T)\\
\end{aligned},
\end{equation}
where $\mathcal{L}_{t2i}$ can be formulated by exchanging $v$ and $t$ in Eq.~(\ref{eq.2}) (\ref{eq.3}), $\epsilon$ is a small number to avoid numerical problems. $q_{i,j}=y_{i,j}/\sum_{k=1}^{N}y_{i,k}$ is the true matching probability, where $y_{i,j}=1$ means that ($v_i$, $t_j$) is a matched pair from the same identity. Eq.~(\ref{eq.3}) shows that the CMPM loss considers many-to-many matching of images and texts under the same identity within a batch. However, due to the mode-seeking nature\footnote[1]{When the true probability density curve exhibits multiple peaks (modes) with areas of zero probability density between them, the approximate probability density curve will be truncated at the points where the true probability density is zero, resulting in the approximation focusing on a specific peak (mode) and disregarding the other peaks (modes).} of the reverse KL divergence $KL(\bm p_i\|\bm q_j)$, the CMPM loss only tries to select a single mode distribution $\bm p_i$ when the true matching distribution $\bm q_i$ of the image $\bm V_i$ has multiple modes in a batch (i.e., there are multiple matching texts)~\cite{CMPM}, which makes the many-to-many correspondences between images and texts under the same identity not fully and effectively utilized.

In general, existing objective functions treat TIReID as a standard image-text matching problem, focusing solely on the one-to-one matching of the data pair ($I_i$, $C_i$) while ignoring the many-to-many matching of images and texts under the same identity. To address this issue, one direct solution is to match multiple positive pairs simultaneously ($(I_i, C_i)$, $(I_j, C_j)$, $(I_i, C_j)$, and $(I_j, C_i)$, where $L_i=L_j$), and we achieve this by contrastive loss~\cite{supcontra}. Figure~\ref{Fig:a} shows the performance of Baselines trained with different losses. The results show that this direct many-to-many matching method by contrastive loss leads to significant performance degradation, we speculate that it may destroy the inherent correspondence of data pairs $(I_i, C_i)$ and $(I_j, C_j)$ with a view due to the vast difference between images and texts under different views. To avoid the direct many-to-many matching way, we deal with this problem from another perspective in the paper. We enrich each single-view text (image) with multiple texts (images) from other views under the same identity to generate a richer text (image) feature. The generated enriched image and text features are aligned in the joint embedding space, indirectly establishing the correspondences between images and texts across views under the same identity.

\subsection{Richer Self}\label{sec3.2}
To enrich each text (image) with information from other views under the same identity, we construct a textual (visual) support set consisting of texts (images) from other views under the same identity and then fuse the text (image) and its corresponding textual (visual) support set to generate a richer textual (visual) feature. Specifically, for text $C_i$, we randomly select $K_t$ texts from the text set of other views under the same identity to form the textual support set $C^s_i=\{C^s_{i,k}\}_{k=1}^{K_t}$. Similarly, for image $I_i$, we also perform similar operations to construct the visual support set $I^s_i=\{I^s_{i,k}\}_{k=1}^{K_v}$. In the following, we design a multi-head attentional fusion (MHAF) module to fusion the feature embeddings of text $C_i$ (image $I_i$) and corresponding support set $C^s_i$ ($I^s_i$). Taking text $C_i$ as an example, we first obtain the feature embeddings ($\bm t_i$ and $\{\bm t^s_{i,k}\}_{k=1}^{K_t}$) of $C_i$ and $C^s_i$ through the textual encoder, and then send them to the MHAF module for feature fusion~\cite{Wsurvey}.
\begin{equation}\
\begin{aligned}
\bm t^r_i=MHAF(\{\bm t_i, \bm t^s_{i,1}, ..., \bm t^s_{i,K_t}\})
\end{aligned}.
\end{equation}
\textbf{Multi-Head Attentional Fusion module}. The MHAF module takes $\bm E=\{\bm t_i, \bm t^s_{i,1}, ..., \bm t^s_{i, K_t}\}\in \mathbb{R}^{(K_t+1)\times d}$ as input and output the enriched textual embedding, which is a weighted sum of all feature embedding in $\bm E$. We employ the multi-head self-attention mechanism to compute the weight.
\begin{equation}\
\begin{aligned}
\bm X_h=\bm E \bm W^X_h, \bm Y_h=\bm E \bm W^Y_h, \bm Z_h=\bm E \bm W^Z_h
\end{aligned},
\end{equation}
\begin{equation}\
\begin{aligned}
% A_h=softmax(\frac{X_hY_h^T}{\sqrt{d}})
\bm A_h=softmax(\bm X_h\bm Y_h^T/\sqrt{d})
\end{aligned},
\end{equation}
where the trainable parameter matrices $\bm W^X_h, \bm W^Y_h, \bm W^Z_h\in \mathbb{R}^{d\times d_c}$, $\bm A_h\in \mathbb{R}^{(K_t+1)\times (K_t+1)}$ is the $h$-th attentional weight matrix ($h=1,2,...,H$), $H$ is the number of multi-head and $d_c=d/H$. Thus, we obtain feature embedding $\hat{\bm E}_h\in \mathbb{R}^{(K_t+1)\times d_c}$ of $h$-th head through
\begin{equation}\
\begin{aligned}
\hat{\bm E}_h=\bm A_h\bm Z_h
\end{aligned}.
\end{equation}
By analogy, the feature embeddings from multiple heads are concatenated to get multi-head embedding $\hat{\bm E}\in \mathbb{R}^{(K_t+1)\times d}$. Finally, we generated the enriched textual embedding $\bm t^r_i\in \mathbb{R}^{d}$ through
\begin{equation}\
\begin{aligned}
\bm t^r_i=MeanPooling(\bm E)+Sum(Fc(\hat{\bm E}))
\end{aligned},
\end{equation}
where $Fc(\cdot)$ represents a fully connected layer with weight $\bm W\in \mathbb{R}^{d\times d}$, $MeanPooling(\cdot)$ and $Sum(\cdot)$ denote the mean pooling operation and summation operation, respectively. Similarly, we can also generate the enriched visual embedding $\bm v^r_i\in \mathbb{R}^{d}$.

Since the generated enriched textual and visual embeddings contain information from multiple views under the same identity, aligning them is equivalent to establishing a many-to-many matching between images and texts of multiple views under the same identity. To align $\bm t^r_i$ and $\bm v^r_i$, we introduce a \textbf{multi-stage CMPM loss} to supervise the learning of the above network. Concretely, for image $I_i$, we can generate multi-stage visual feature set $\{\bm v^l_i, \bm v^h_i, \bm v^r_i\}$, where $v^l_i\in \mathbb{R}^{d_1}$ and $v^h_i\in \mathbb{R}^{d}$ (i.e., $\bm v_i$ above) are the features generated by the 3rd and 4th residual blocks of the visual encoder, namely ResNet50. Similarly, for text $C_i$, we can also generate a set of multi-stage textual features $\{\bm t^l_i, \bm t^h_i, \bm t^r_i\}$, where $\bm t^l_i\in \mathbb{R}^{d_1}$ and $\bm t^h_i\in \mathbb{R}^{d}$ (i.e., $\bm t_i$ above) are the features generated by the 1$\times$1 convolutional block and residual block of the TextCNN~\cite{tipcb} network, respectively.

For convenience, let $\bm V^l\in \mathbb{R}^{N\times d_1}$, and $\bm V^h, \bm V^r\in \mathbb{R}^{N\times d}$ be matrices that consist of a batch of visual embeddings from multiple stages, respectively. Let $\bm T^l\in \mathbb{R}^{N\times d_1}$, and $\bm T^h, \bm T^r\in \mathbb{R}^{N\times d}$ be matrices that consist of a batch of textual embeddings from multiple stages, respectively. The multi-stage CMPM loss is defined as
\begin{equation}\
\begin{aligned}
\mathcal{L}_{ms}=\mathcal{L}^l_{cmpm}(\bm V^l, \bm T^l)+\mathcal{L}^h_{cmpm}(\bm V^h, \bm T^h)+\mathcal{L}^r_{cmpm}(\bm V^r, \bm T^r)
\end{aligned}.
\end{equation}

Furthermore, in order to ensure that the MHAF module aggregates as much information from other views as possible while preserving information from the current view, we design a \textbf{cross-stage CMPM loss}, which is defined as follows 
\begin{equation}\
\begin{aligned}
\mathcal{L}_{cs}=\mathcal{L}^{h\rightarrow r}_{cmpm}(\bm V^h, \bm T^r)+\mathcal{L}^{r\rightarrow h}_{cmpm}(\bm V^r, \bm T^h)
\end{aligned}.
\end{equation}

The overall optimization objective is defined as: 
\begin{equation}\
\begin{aligned}
\mathcal{L}_{teacher}=\mathcal{L}_{ms}+\lambda_1\mathcal{L}_{cs}
\end{aligned},
\end{equation}
where $\lambda_1$ is a hyper-parameter to control the importance of $\mathcal{L}_{cs}$.

Based on the above model, we can effectively model the many-to-many matching between images and texts under the same identity. However, the model requires additional images and texts from other views under the same identity, which are not available during inference. In inference, we can only match a single-view text to each image in the candidate pool. Therefore, we utilize the above model as the teacher model and introduce knowledge distillation to train a simple and efficient model (student model) for inference with a single text/image as input. Since the student and teacher models transfer knowledge between the same identities (self-transfer), we call the teacher model with richer knowledge the "Richer Self".

\subsection{Distilling "Richer" Knowledge}\label{sec3.3}
The student network can be any simple and basic dual encoding network. In the work, we keep the same structure as the teacher network with the MHAF module removed, which takes a single text/image as input. The teacher model focuses on fusing information from multiple views and learning multi-view associations to better model many-to-many matching relationships. To transfer this powerful ability for inference, we distill the richer knowledge of the teacher network to the student network. By doing so, we expect the student network to acquire the ability to multi-view semantic association and reasoning based on only a single input.

The training of the student network is supervised by two parts: (1) Supervised by the basic cross-modal matching loss so that it has the basic modality alignment ability. (2) Supervised by the teacher network via knowledge distillation to transfer rich knowledge to the student network. Formally, for a batch of $N$ paired image-text tuples $\{I_i, C_i\}_{i=1}^{N}$, we can get visual and textual embeddings $\bm V_s^l\in \mathbb{R}^{N\times d_1}$, $\bm V_s^h\in \mathbb{R}^{N\times d}$, $\bm T_s^l\in \mathbb{R}^{N\times d_1}$, $\bm T_s^h\in \mathbb{R}^{N\times d}$ from multiple stages by the student model, respectively. 

\noindent\textbf{Cross-modal Matching.} Similarly, we also employ the multi-stage CMPM loss to supervise the student model.
\begin{equation}\
\begin{aligned}
\mathcal{L}^s_{ms}=\mathcal{L}^l_{cmpm}(\bm V_s^l, \bm T_s^l)+\mathcal{L}^h_{cmpm}(\bm V_s^h, \bm T_s^h)
\end{aligned}.
\end{equation}

To better empower the student network with the ability of multi-view semantic association and reasoning, we utilize the intra-modal features and inter-modal semantic relations of the teacher network as supervision signals to supervise the student network.

\noindent\textbf{Intra-modal Feature Distillation.} We first transfer knowledge by enforcing the student model to mimic the enriched features output by the teacher model, which is formulated to minimize the mean square error (MSE) between the output features of teacher and student networks.
\begin{equation}\
\begin{aligned}
\mathcal{L}_{KD-F}=\underbrace{MSE(\bm V_s^h, \bm V^r)}_{\mathcal{L}_{KD-F_v}}+\underbrace{MSE(\bm T_s^h, \bm T^r)}_{\mathcal{L}_{KD-F_t}}
\end{aligned}.
\end{equation}

\noindent\textbf{Inter-modal Relation Distillation.} To propagate the inter-modal relation of the teacher model to the student model, we compute the inter-modal similarity matrices as $\bm S_t=\bm V^r(\bm T^r)^T\in \mathbb{R}^{N\times N}$ and $\bm S_s=\bm V_s^h(\bm T_s^h)^T\in \mathbb{R}^{N\times N}$ for teacher and student networks. The inter-modal relation distillation loss is formulated as
\begin{equation}\
\begin{aligned}
\mathcal{L}_{KD-R}=\frac{1}{N}\|\bm S_s-\bm S_t\|_F^2
\end{aligned},
\end{equation}
where $\|\cdot\|_F$ denotes Frobenius norm. Integrating the above losses, the objection function $\mathcal{L}_{student}$ for the student model is as follows
\begin{equation}\
\begin{aligned}
\mathcal{L}_{student}=\lambda_2\mathcal{L}^s_{ms}+\lambda_3(\mathcal{L}_{KD-F}+\mathcal{L}_{KD-R})
\end{aligned},
\end{equation}
where $\lambda_2$ and $\lambda_3$ balance the focus on different loss terms.

\noindent\textbf{Inference}. Note that only the student model is used for inference since the support set is inaccessible during inference. During inference, we first generate textual and visual features for the text query and image candidate using the student network, then calculate the cosine similarity between them.

\section{Experiments}
We comprehensively validate the performance of LCR$^2$S on several public datasets. In the following subsections, we first introduce the datasets and metrics used in the experiments, as well as the implementation details. We then showcase the overall performance of our LCR$^2$S and compare it with state-of-the-art methods on each dataset. Finally, we conduct ablation studies to assess the effectiveness of each component of our method.

\vspace{-6.5pt}
\subsection{Datasets and Metrics}
\textbf{CUHK-PEDES}~\cite{GNA} contains 40,206 images and 80,412 text descriptions of 13,003 persons, each image is manually annotated with 2 descriptions, and the average length of each description is no less than 23 words. Following~\cite{GNA}, we train our model on the training set of 34,054 images and 68,108 descriptions of 11,003 persons, and report results on the test set of 3,074 images and 6,148 descriptions of 1000 persons. 

\noindent\textbf{ICFG-PEDES}~\cite{SSAN} consists of 54522 image-text pairs for 4,102 persons, with each text description having an average length of 37 words. Following~\cite{SSAN}, we use the standard split of 34674 image-text pairs of 3102 persons, and 19848 image-text pairs of the remaining 1000 persons for training, and testing. 

\noindent\textbf{RSTPReid}~\cite{DSSL} contains 41010 textual descriptions and 20505 images of 4101 persons, each of which contains 5 images captured by 15 cameras, and each image corresponds to 2 text descriptions with a length of no less than 23 words. Following~\cite{DSSL}, we split the dataset into 3701, 200, and 200 persons for training, validation, and testing.

\noindent\textbf{Metrics}. We evaluate the retrieval performance using Rank-K accuracy (K=1, 5, 10), which represents the percentage of queries that retrieve at least one ground truth among the top K results.

\vspace{-6.5pt}
\subsection{Implementation Details}
We conduct the experiments on the PyTorch with a single RTX3090 24GB GPU. The teacher model includes a visual encoder, a textual encoder, and an MHAF module, where the visual and textual encoders are kept consistent with~\cite{tipcb}. While the student model is a basic dual encoding network that only contains the same visual and textual encoders as the teacher model. All input images are resized to 384$\times$128, and the maximum length of text sequences is set to 64. Random horizontal flipping and random crop with padding are used for image augmentation. The feature embedding dimensions are set to $d_1=1024$ and $d=2048$. For MHAF, we set the number of multi-head $H$ to 16, and each text (image) has a textual (visual) support set consisting of $K_t=1$ ($K_v=1$) other texts (images) under the same identity. The hyperparameters for balancing multiple losses $\lambda_1$, $\lambda_2$, and $\lambda_3$ are set to 1, 0.9, and 1, respectively. We train our model using Adam optimizer with a batch size of 64 and adopt a linear warmup strategy. During training, we employed a staged training strategy. Specifically, we first train the teacher model for 60 epochs with a learning rate initialized to 1e-3, which is then decreased by 0.1 at the 30th, 40th, and 50th epoch, respectively. After that, we freeze the teacher model and train the student model from scratch for 60 epochs. For the student model, we set different modules with different initial learning rates, where the visual encoder is set to 1e-4, the others are set to 1e-3, and the learning rate is decreased by a factor of 0.1 at the 30th, and 45th epoch, respectively.

\begin{table}[t!]\small
\setlength{\abovecaptionskip}{0.1cm} 
\setlength{\belowcaptionskip}{-0.2cm}
\centering {\caption{Performance comparison with state-of-the-art methods on CUHK-PEDES. '-' denotes that no reported result is available.}\label{Tab:1}
\renewcommand\arraystretch{1.1}
\setlength{\tabcolsep}{2.0mm}{
\begin{tabular}{c|c|cccc}
%{m{2.5cm}<{\centering}|m{1.5cm}<{\centering}|m{1.0cm}<{\centering}m{1.0cm}<{\centering}m{1.0cm}<{\centering}}
\hline
%\multicolumn{5}{|c|}{Flower}\\
 \hline
  Methods  &Ref & Rank-1 & Rank-5 & Rank-10 & mAP\\
  \hline
  MCCL~\cite{MCCL}  & ICASSP19 & 50.58 & - & 79.06 & - \\

  A-GANet~\cite{A-GANet}  & MM19 & 53.14 & 74.03 & 81.95 & - \\

  TIMAM~\cite{TIMAM}  & ICCV19 & 54.51 & 77.56 & 84.78 & - \\

  MIA~\cite{MIA}  & TIP20 & 53.10 & 75.00 & 82.90 & - \\
  
  PMA~\cite{PMA}  & AAAI20 & 53.81 & 73.54 & 81.23 & - \\

  TDE~\cite{TDE}  & MM20 & 55.25 & 77.46 & 84.56 & - \\

  ViTAA~\cite{vitaa}  & ECCV20 & 55.97 & 75.84 & 83.52 & - \\

  IMG-Net~\cite{IMG-Net}  & JEI20 & 56.48 & 76.89 & 85.01 & - \\

  CMAAM~\cite{CMAAM}  & WACV20 & 56.68 & 77.18 & 84.86 & - \\

  HGAN~\cite{HGAN}  & MM20 & 59.00 & 79.49 & 86.62 & 37.80 \\

  CMKA~\cite{CMKA}   & TIP21 & 54.69  &73.65  &81.86 & - \\

  DSSL~\cite{DSSL}   & MM21 & 59.98  &80.41  &87.56 & - \\
  
  MGEL~\cite{mgel}   & IJCAI21 & 60.27  &80.01  &86.74 & - \\

  SSAN~\cite{SSAN}   & arXiv21 & 61.37  &80.15  &86.73 & - \\
  
  LapsCore~\cite{LapsCore}  & ICCV21 & 63.40 & - & 87.80 & - \\

  TextReID~\cite{TextReID}   & BMVC21 & 64.08  & 81.73  & 88.19 & \textbf{60.08} \\

  SUM~\cite{sum}  & KBS22 & 59.22  & 80.35  & 87.60 & 37.91 \\

  ACSA~\cite{ACSA}  & TMM22  & 63.56   & 81.40  & 87.70 & - \\

  MANet~\cite{MANet}  & arXiv22  & 63.92 & 82.15 & 87.69 & -  \\
  
  IVT~\cite{ivt}   & ECCVW22 & 64.00  & 82.72   & 88.95 & 58.99 \\

  SRCF~\cite{SRCF} & ECCV22 & 64.04   & 82.99   & 88.81  & - \\

  LBUL~\cite{LBUL} & MM22 & 64.04  & 82.66  & 87.22  & - \\
  
  SAF~\cite{saf}   & ICASSP22 & 64.13  & 82.62  & 88.40 & - \\

  TIPCB~\cite{tipcb}   & Neuro22 & 64.26  & 83.19  & 89.10 & - \\
  
  CAIBC~\cite{caibc}   & MM22 & 64.43  & 82.87   & 88.37 & - \\
  
  AXM-Net~\cite{AXM-Net}   & AAAI22 & 64.44  & 80.52  & 86.77 & 58.73 \\ 

  C$_2$A$_2$~\cite{C2A2}   & MM22 & 64.82  & \underline{83.54}  & \textbf{89.77} & - \\ 

  LGUR~\cite{lgur}    & MM22  & \underline{65.25}  & 83.12  & 89.00  & - \\

  RKT~\cite{RKT}   & TMM23 & 61.48  &80.74  &87.28  & - \\
  \hline
  \textbf{LCR$^2$S}  & MM23 & \textbf{67.36}  &  \textbf{84.19}  &  \underline{89.62} & \underline{59.24}\\
  \hline\hline
\end{tabular}}}
\vspace{-0.3cm}
\end{table}

\vspace{-4pt}
\subsection{Comparison with State-of-the-Art Methods}
In this section, we present the quantitative results of our LCR$^2$S and compare them with existing TIReID methods on different datasets. Tables~\ref{Tab:1},~\ref{Tab:2},~\ref{Tab:3} present the results on CHUK-PEDES, ICFG-PEDES, and RSTPReid. It is evident that LCR$^2$S outperforms all the comparison methods on the three datasets, especially on CHUK-PEDES and ICFG-PEDES by a clear margin. Specifically, for CHUK-PEDES, LCR$^2$S achieves 67.36\%, 84.19\% and 89.62\% on Rank-1, Rank-5 and Rank-10, which have improvements of 2.11\%, 1.07\%, and 0.62\% on these metrics compared to the recent state-of-the-art method LGUR~\cite{lgur}. The accuracy at Rank-1 on ICFG-PEDES and RSTPReid is 57.83\% and 54.95\%, which is improved by 0.51\% and 3.4\% over the current state-of-the-art methods LGUR~\cite{lgur} and C$_2$A$_2$~\cite{C2A2}, respectively. The current SOTA methods~\cite{lgur, SRCF} require a separate local branch to extract fine-grained part-level visual and textual features for retrieval except for the modality-specific encoder, which results in higher computational cost and slower retrieval speed. In contrast, our method only uses a basic dual encoding network for inference, consisting of visual and textual encoders. This means LCR$^2$S can achieve higher retrieval efficiency and improve the performance without additional cost at inference. Our LCR$^2$S consistently achieves new state-of-the-art performance on all three popular datasets, demonstrating its effectiveness and superiority. The reason for its simplicity and effectiveness is that it addresses the fundamental problem of TIReID, which is many-to-many matching.

\begin{table}[!t]\small
\setlength{\abovecaptionskip}{0.1cm}
\centering {\caption{Performance comparison with state-of-the-art methods on ICFG-PEDES.}\label{Tab:2}
\renewcommand\arraystretch{1.1}
\setlength{\tabcolsep}{2.0mm}{
\begin{tabular}{c|c|cccc}
\hline
%\multicolumn{5}{|c|}{Flower}\\
 \hline
  Methods & Ref & Rank-1 & Rank-5 & Rank-10 & mAP\\
  \hline
  CMPM/C~\cite{CMPM} & ECCV18 & 43.51 & 65.44 & 74.26 & - \\

  SCAN~\cite{SCAN} & ECCV18 & 50.05 & 69.65 & 77.21 & - \\

  Dual Path~\cite{Dual} & TOMM20 & 38.99 & 59.44 & 68.41 & - \\

  MIA~\cite{MIA} & TIP20 & 46.49 & 67.14 & 75.18 & - \\

  ViTAA~\cite{vitaa} & ECCV20 & 50.98 & 68.79 & 75.78 & - \\

  SSAN~\cite{SSAN} & arXiv21 & 54.23 & 72.63 & 79.53 & - \\

  TIPCB~\cite{tipcb}   & Neuro22  & 54.96  & 74.72   & \underline{81.89} & - \\
  
  IVT~\cite{ivt}   & ECCVW22 & 56.04  & 73.60  & 80.22 & - \\

  SRCF~\cite{SRCF} & ECCV22 & 57.18   & \underline{75.01}  & 81.49 & - \\

  LGUR~\cite{lgur} & MM22  & \underline{57.42} & 74.97  & 81.45 & - \\
  \hline
  \textbf{LCR$^2$S}  & MM23 & \textbf{57.93}  & \textbf{76.08}  & \textbf{82.40} & \textbf{38.21} \\
  \hline\hline
\end{tabular}}}
\vspace{-0.25cm}
\end{table}

\begin{table}[!t]\small
\setlength{\abovecaptionskip}{0.1cm}
\centering {\caption{Performance comparison with state-of-the-art methods on RSTPReid.}\label{Tab:3}
\renewcommand\arraystretch{1.1}
\setlength{\tabcolsep}{2.0mm}{
\begin{tabular}{c|c|cccc}
\hline
%\multicolumn{5}{|c|}{Flower}\\
 \hline
  Methods & Ref & Rank-1 & Rank-5 & Rank-10 & mAP \\
  \hline
  IMG-Net~\cite{IMG-Net}  & JEI20 & 37.60 & 61.15 & 73.55 & - \\

  AMEN~\cite{amen}   & PRCV21  & 38.45  & 62.40  & 73.80 & - \\

  DSSL~\cite{DSSL}   & MM21 & 39.05  &62.60  &73.95 & - \\

  SSAN~\cite{SSAN} & arXiv21 & 43.50 & 67.80 & 77.15 & - \\

  SUM~\cite{sum}   & KBS22 & 41.38  & 67.48  & 76.48 & - \\

  LBUL~\cite{LBUL}  & MM22   & 45.55  & 68.20 & 77.85 & - \\
  
  IVT~\cite{ivt}   & ECCVW22 & 46.70  & 70.00  & 78.80 & - \\

  ACSA~\cite{ACSA}  & TMM22  & 48.40  & 71.85  & 81.45 & - \\
  
  C$_2$A$_2$~\cite{C2A2}   & MM22 & \underline{51.55}  & \textbf{76.75} &  \textbf{85.15} & - \\ 
  \hline
  \textbf{LCR$^2$S}  & MM23 & \textbf{54.95}  & \underline{76.65}  & \underline{84.70} & \textbf{40.92} \\
  \hline\hline
\end{tabular}}}
\vspace{-0.3cm}
\end{table}

\vspace{-4pt}
\subsection{Ablation Study}
To assess the effectiveness of each component in LCR$^2$S, we conduct a comprehensive set of ablation experiments, all under the same experimental settings. "Baseline" represents the student network trained only by the basic cross-modal matching loss.

\noindent\textbf{Distillation strategy.} The distillation strategy for training the student model in LCR$^2$S is crucial as it endows the student network with the ability to multi-view semantic association and reasoning. Table~\ref{Tab:4} reports the effect of different distillation strategies. The results show that even distilling knowledge from a single modality can lead to significant improvements. This proves that it is unreliable to match a single-view text with images from multiple views due to the vast variation of images and texts in different views. The 4th, 5th, 6th, and 8th rows show that the inter-modal relation distillation can further improve the performance. The results in the 4th row show that transferring knowledge only by inter-modal relation distillation loss can outperform all compared methods in Table~\ref{Tab:1}. This confirms the importance of the inter-modal relation distillation loss for the student network to master multi-view semantic association and reasoning abilities. The best performance is achieved when knowledge of both modalities is distilled simultaneously.

\noindent\textbf{Fusion strategy.} In LCR$^2$S, we use the modality-shared MHAF module to fuse the modality-specific feature with its corresponding support set. To validate the effectiveness of MHAF, we compare three fusion schemes by replacing MHAF with Mean Pooling, Cross-attention~\cite{LINAS}, modality-specific MHAF (w/o Shared). The performance with specific feature fusion blocks is reported in Table~\ref{Tab:5}, which shows the superiority of MHAF. The number of multi-head $H$ in MHAF is also a parameter that significantly affects performance. Figure~\ref{Fig:3} (top) shows that as the number of multi-head increases, the performance improves compared to when $H$=1, which highlights the importance of multi-head. The best retrieval performance is achieved when $H$=16.

\begin{table}[!t]\small
\setlength{\abovecaptionskip}{0.1cm}
\centering {\caption{Ablation studies of distillation strategy on CUHK-PEDES.}\label{Tab:4}
\renewcommand\arraystretch{1.1}
\setlength{\tabcolsep}{1.0mm}{
\begin{tabular}{c|ccc|ccc}
\hline
\hline
Methods  &  $\mathcal{L}_{KD-F_t}$  & $\mathcal{L}_{KD-F_v}$ & $\mathcal{L}_{KD-R}$ & Rank-1 & Rank-5 & Rank-10\\
\hline
Baseline  &   &   &   & 62.31  &  82.29  &  88.76 \\
+T     &\checkmark &   &   &  63.29  &  82.84  &  89.26  \\
+I     &   &\checkmark  &  &  63.79  &  83.23  &  89.61  \\
+R     &   &  &\checkmark  &  65.84  &  84.30  &  89.74  \\
+TR    &\checkmark &  & \checkmark & 66.22  &  83.57  &  89.25  \\
+IR    &   &\checkmark  & \checkmark & 66.30  &  83.65  &  89.54  \\
+TI    &\checkmark  & \checkmark &   &  64.69  &  83.33  &  89.61 \\
+TIR (LCR$^2$S)  &\checkmark  & \checkmark & \checkmark & 67.36  &  84.19  &  89.62  \\
\hline\hline
\end{tabular}}}
\vspace{-0.3cm}
\end{table}

\begin{table}[!t]\small
\setlength{\abovecaptionskip}{0.1cm}
\centering {\caption{Effects of different feature fusion strategies on CUHK-PEDES.}\label{Tab:5}
\renewcommand\arraystretch{1.1}
\setlength{\tabcolsep}{2.6mm}{
\begin{tabular}{c|ccc}
\hline
 \hline
  Method  & Rank-1 & Rank-5 & Rank-10 \\
  \hline
  Mean Pooling   &  66.69  &  83.90  &  89.79 \\

  Cross-attention~\cite{LINAS}  &  66.31  &  83.62  &  89.82 \\

  w/o Shared  &  66.20  &  84.17  &  89.74  \\

  MHAF (Ours) &  67.36  &  84.19  &  89.62   \\
  \hline\hline
\end{tabular}}}
\vspace{-0.3cm}
\end{table}

\begin{table}[!t]\small
\setlength{\abovecaptionskip}{0.1cm}
\centering {\caption{Ablation studies of teacher loss function on CUHK-PEDES.}\label{Tab:6}
\renewcommand\arraystretch{1.1}
\setlength{\tabcolsep}{0.8mm}{
\begin{tabular}{ccccc|ccc}
\hline
\hline
$\mathcal{L}^r_{cmpm}$ & $\mathcal{L}^l_{cmpm}$  & $\mathcal{L}^h_{cmpm}$  & $\mathcal{L}^{h\rightarrow r}_{cmpm}$ & $\mathcal{L}^{r\rightarrow h}_{cmpm}$ & Rank-1 & Rank-5 & Rank-10\\
\hline
\checkmark  &             &   &   &   &  64.69  &  82.68  &  88.39  \\
\checkmark  &\checkmark   &   &   &   &  65.55 &  83.07  &  88.55 \\
\checkmark  &\checkmark   &\checkmark &  &  & 66.12  &  83.67  &  89.75  \\
\checkmark  &\checkmark   &\checkmark &\checkmark  &  &  66.87 & 83.91  &  89.59  \\
\checkmark  &\checkmark   &\checkmark &  & \checkmark &  66.32  &  84.16  &  89.48  \\
\checkmark  &\checkmark   &\checkmark & \checkmark & \checkmark & 67.36  &  84.19  &  89.62  \\
\hline\hline
\end{tabular}}}
\vspace{-0.3cm}
\end{table}

\noindent\textbf{Teacher Loss.} To align the enriched visual and textual features and establish many-to-many correspondences, we employ five alignment losses. Extensive ablation experiments are conducted on CUHK-PEDES to validate their effectiveness and the results are shown in Table~\ref{Tab:6}. We observe that the multi-stage matching loss $\mathcal{L}_{ms}$ (the 3rd row) can lead to a larger performance gain (1.43\% improvement in Rank-1) compared to the base loss $\mathcal{L}^r_{cmpm}$. This confirms the effectiveness of the shallow-to-deep alignment strategy for cross-modal alignment. Moreover, the combination of the multi-stage and cross-stage matching losses results in a 1.24\% improvement in Rank-1, which reveals the effect of $\mathcal{L}_{ms-cm}$. This cross-stage matching loss $\mathcal{L}_{ms-cm}$ not only helps cross-modal alignment but also ensures the effectiveness of MHAF in fusing features.

\begin{figure}[t!]
\setlength{\abovecaptionskip}{0.1cm}
\setlength{\belowcaptionskip}{-0.3cm}
\centering
\includegraphics[width=\linewidth]{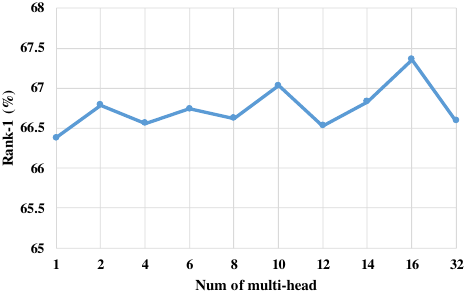}
\includegraphics[width=\linewidth]{SupportSize.pdf}
\caption{Effects of important parameters, (a) the number of multi-head in MHAF; (b) support set size on CUHK-PEDES.}
\label{Fig:3}
\vspace{-0.3cm}
\end{figure}

\noindent\textbf{Necessity of KD.} To validate the necessity of the knowledge distillation mechanism, we include additional experimental results showcasing direct inference through the teacher network without the MHAF module. As depicted in Table~\ref{Tab:7}, the retrieval performance of the teacher network is even inferior to that of the
baseline model. The baseline model only contains images and text backbones, while the teacher network additionally introduces an MHAF module. The backbones primarily focus on modeling one-to-one matching, while the MHAF module is responsible for fusing information from multiple views to model the many-to-many matching. due to the difference in task focus between the backbone and the MHAF module, the one-to-one matching ability of the backbone will be interfered by the MHAF module, resulting in even poorer performance compared to the baseline. 

And the introduction of the cross-stage CMPM loss ($L_{cs}$) further strengthens the interference of the MHAF module on the backbone. We conducted some experiments to validate this observation. Table~\ref{Tab:8} presents the results for different variants of the teacher network. The results in the second row show that the introduction of the MHAF module reduces the backbone's one-to-one matching ability, leading to performance degradation. We introduce the cross-stage CMPM loss to interact between single-view features and multi-view features, which is equivalent to enhancing the interaction between the backbone and MHAF modules, further increasing the interference between them, and resulting in additional performance degradation. Additionally, when we further increase the interaction between modules by narrowing the distance between the inter-modal single-view and multi-view feature similarity matrix ($L_{cr}$), the performance significantly drops. These results strongly support our previous statement. Note that the one-to-one matching ability of the teacher network is not the primary focus of our attention. Our main emphasis lies in the effective integration of multi-view information and the modeling of many-to-many matching. While the introduction of the cross-stage CMPM loss is not beneficial for the backbone's one-to-one matching ability, it effectively promotes the MHAF module's ability to fuse multi-view information and model many-to-many matching (which has been demonstrated in Table~\ref{Tab:6}). This aligns with our intended purpose for the teacher network.

\begin{table}[t!]\small
\setlength{\abovecaptionskip}{0.1cm}
\centering {\caption{Ablation studies of the necessity of KD on CUHK-PEDES.}\label{Tab:7}
\renewcommand\arraystretch{1.1}
\setlength{\tabcolsep}{2.6mm}{
\begin{tabular}{c|cccc}
\hline
\hline
  Methods  & Rank-1 & Rank-5 & Rank-10 & mAP \\
  \hline
  Baseline & 62.31 &  82.29 &  88.76 & 52.46  \\  
  Teacher model  & 61.53 & 81.65 & 87.61 & 52.13  \\
  Student model & 67.36 & 84.19 & 89.62 & 59.24  \\
  \hline
  \hline
\end{tabular}}}
\vspace{-0.3cm}
\end{table}

\begin{table}[t!]\small
\setlength{\abovecaptionskip}{0.1cm}
\centering {\caption{Ablation studies of the interference of the MHAF on the backbone in the teacher network on CUHK-PEDES.}\label{Tab:8}
\renewcommand\arraystretch{1.1}
\setlength{\tabcolsep}{2.6mm}{
\begin{tabular}{c|cccc}
\hline
\hline
  Methods  & Rank-1 & Rank-5 & Rank-10 & mAP \\
  \hline
  Baseline & 62.31 &  82.29 &  88.76 & 52.46  \\ 
  +MHAF & 61.94 & 81.91 & 88.45 & 51.91  \\
  +MHAF+$L_{cs}$ & 61.53 & 81.65 & 87.61 & 52.13  \\
  +MHAF+$L_{cs}$ +$L_{cr}$ & 58.02 & 79.7 & 86.74 & 49.5 \\
  \hline
  \hline
\end{tabular}}}
\vspace{-0.3cm}
\end{table}

\begin{table}[t!]\small
\setlength{\abovecaptionskip}{0.1cm}
\centering {\caption{Ablation studies of the importance of multi-view information on CUHK-PEDES.}\label{Tab:9}
\renewcommand\arraystretch{1.1}
\setlength{\tabcolsep}{2.6mm}{
\begin{tabular}{c|cccc}
\hline
\hline
  Methods  & Rank-1 & Rank-5 & Rank-10 & mAP \\
  \hline
  Baseline & 62.31 &  82.29 &  88.76 & 52.46   \\ 
  Setting~1 & 63.18 & 82.95 & 89.18 & 53.22  \\
  Setting~2 & 63.62 & 83.36 & 89.30 & 53.91 \\
  LCR$^2$S & 67.36 & 84.19 & 89.62 & 59.24 \\
  \hline
  \hline
\end{tabular}}}
\vspace{-0.3cm}
\end{table}
\noindent\textbf{Importance of Multi-view Information.} We make several additions to our experiments to further demonstrate the effectiveness of introducing multi-view information and modeling the many-to-many matching. In the first set of experiments (Setting 1), we utilized a trained Baseline network (with the same structure as the student network but without the MHAF module) as the teacher network to transfer knowledge to the student network. Similarly, in the second set of experiments (Setting 2), we maintained the same structure as the current teacher network with the MHAF module. However, since the MHAF module requires at least two features for fusion, we duplicated the single-view features and fed them into the MHAF module together. Note that multi-view information was not introduced in either of these experiment sets, and the results are summarized in Table~\ref{Tab:9}. Despite the absence of additional multi-view information in the teacher network, the student network exhibited noticeable improvement in both settings, benefiting from the stronger supervision signal provided. When we transitioned from single-view to multi-view inputs, even with the introduction of just one additional view, we observed a significant performance boost. Compared to the previous two settings, the Rank-1 accuracy showed a remarkable improvement of 4.18\% and 3.74\%, respectively. This clearly validates the value and potential of introducing multi-view information and many-to-many matching relationship in TIReID. 

\noindent\textbf{Support Set Size.} Support set sizes $K_t$ and $K_v$ are crucial parameters for learning enriched features. Each identity has multiple images and multiple texts from multiple views. To investigate the impact of the support set size, for each text (image), we randomly select a different number of texts (images) except itself from multi-view text (image) set of the same identity to form the textual (visual) support set. Figure~\ref{Fig:3} (bottom) illustrates the results of various support set sizes. We observe that the retrieval performance is better when $K_t\leq3$ and $K_v\leq2$. As images and texts differ significantly under different views and contain some pedestrian-independent noise, a large support set size may introduce too much noise, making it challenging for the model to learn effective many-to-many relationships, and the model may not converge easily. For computational efficiency, we set $K_t$=1 and $K_v$=1 in the experiment.

\noindent\textbf{Computational Complexity.} We analyze the model complexity and compare our method with several representative TIReID methods. The findings are summarized in Table~\ref{Tab:10}, reporting the number of model parameters (Params), the floating-point operations required per input image-text pair (FLOPs) during training, and the retrieval time (Time) at the inference stage. The introduction of the teacher network contributes to the overall complexity of our model. However, it is crucial to note that the teacher network serves a role similar to pre-training and is solely utilized as a supervision signal to guide the training of the student network, and it is not employed during inference. The student network used for inference serves as a basic baseline network and only consists of the necessary image and text backbones without introducing any additional modules. Table~\ref{Tab:9} reveals that our student network shares the same computational complexity as the baseline. Regarding the teacher network, in addition to incorporating necessary backbones, it introduces a feature fusion module to effectively integrate multi-view information. While this incurs an additional computational cost, the resulting performance gain is substantial. Notably, the table demonstrates that our method exhibits a clear advantage in terms of inference efficiency when compared to other methods, further validating the practicality of our method.
\begin{table}[t!]\small
\setlength{\abovecaptionskip}{0.1cm}
\centering {\caption{Comparison of computational complexity and retrieval time on CUHK-PEDES of Several Methods.}\label{Tab:10}
\renewcommand\arraystretch{1.1}
\setlength{\tabcolsep}{2.6mm}{
\begin{tabular}{c|cccc}
\hline
\hline
  Methods  & Params & FLOPs & Time & Rank-1 \\
  \hline
  Baseline & 144.04M & 12.37 & 17.75s & 62.31   \\ 
  Teacher Model & 160.82M & 24.80 & - & -  \\
  SSAN~[1] & 97.86M & 18.14 & 21.36s & 61.37\\
  TIPCB~[2] & 184.75M & 43.86 & 25.04s & 64.26 \\
  Student Model & 144.04M & 12.37 & 17.75s & 67.36 \\
  \hline
  \hline
\end{tabular}}}
\vspace{-0.3cm}
\end{table}

\begin{figure}[t!]
\setlength{\abovecaptionskip}{0.1cm}
  \centering
  \includegraphics[width=\linewidth]{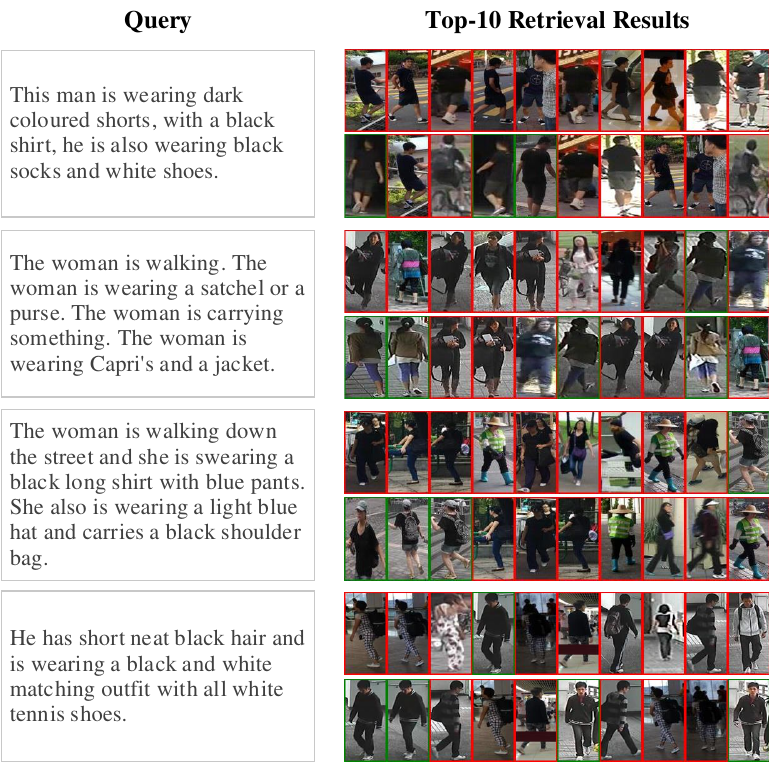}
  \caption{Some examples of text-to-image retrieval to compare Baseline (the 1st row) with LCR$^2$S (the 2nd row) for each text query on CUHK-PEDES. Matched and mismatched images are marked with green and red rectangles, respectively.}
  \label{Fig:4}
  \vspace{-0.3cm}
\end{figure}

\noindent\textbf{Some Retrieval Examples.} In Figure~\ref{Fig:4}, we show a comparison of top-10 retrieval results (our LCR$^2$S versus Baseline) on CUHK-PEDES. As shown, LCR$^2$S achieves more accurate retrieval results in cases where Baseline retrieval fails. The difference between the student model used for inference in LCR$^2$S and Baseline is the additional supervision signal from the teacher network during training. Through the supervision of the teacher network, the student network for inference gains the ability to multi-view semantic association and reasoning, which enables it to accurately retrieve images from multi-view under the same identity with a single text containing multi-view information.

\noindent\textbf{Limitations.} Appropriately larger support sets should lead to greater performance gains, but the results show a sharp drop when $K_t\textgreater3$ and $K_v\textgreater2$. We conjecture that this is caused by introducing too much modality-specific noise, and we believe that a "suppression follow by fusion" may be an effective solution. While LCR$^2$S is simple and effective, its training is computationally expensive due to the additional support set required. Moreover, we indirectly consider many-to-many matching under the same identity from another perspective in the paper. We plan to directly design the loss function for effective many-to-many matching in future work.

\section{Conclusion}
In this paper, we propose a Learning Comprehensive Representations with Richer Self framework (LCR$^2$S), a simple yet effective teacher-student structure designed to mine many-to-many correspondences between multiple image-text pairs across views under the same identity from a novel perspective for TIReID. The teacher network which takes text/image and its corresponding support set as input is designed to fuse multi-view information to generate richer text/image embeddings, followed by aligning them to model many-to-many matching. And we introduce a simple and lightweight student network with a single text/image as input for inference, which inherits the ability of the teacher network through knowledge distillation. Thus, the student model can generate a comprehensive representation containing multi-view information with only a single-view input to perform accurate text-to-image retrieval. Significant performance gains and extensive ablation results on three public TIReID benchmarks prove the superiority and effectiveness of our proposed LCR$^2$S. Note that LCR$^2$S is model-agnostic and can be applied to any dual encoding network.

% \begin{acks}
% This work was supported in part by the National Key Research and Development Program of China under Grant 2018AAA0102002, the National Natural Science Foundation of China under Grant 62172212.
% \end{acks}

% \clearpage
\bibliographystyle{ACM-Reference-Format}
\bibliography{sample-base}


\end{document}